\documentclass[runningheads]{llncs}
\usepackage{graphicx}
\usepackage{amsmath}
\usepackage{amsfonts} 
\usepackage{amssymb}
\usepackage{booktabs}
\usepackage{graphicx}
\usepackage{multirow}
\usepackage{multicol}
\usepackage{enumitem}

\usepackage{latexsym}
\usepackage{wasysym}
\usepackage{marvosym}
\usepackage{ifsym}
\usepackage{xcolor}

\begin{document}
\title{Secure and Scalable Face Retrieval \\via Cancelable Product Quantization*}

\author{Haomiao Tang\inst{1,\dag, \ddag} \and
Wenjie Li\inst{1, \dag} \and
Yixiang Qiu\inst{1, \ddag} \\
Genping Wang\inst{1, 2(\textrm{\Letter})} \and
Shu-Tao Xia\inst{1}
}

\authorrunning{H. Tang and W. Li et al.}
%
\institute{Shenzhen International Graduate School, Tsinghua University, Shenzhen, China \\
\email{\{thm23, liwj20, qiu-yx24\}@mails.tsinghua.edu.cn, xiast@sz.tsinghua.edu.cn}
\and Qiji Technology, Shenzhen, China\\
\email{genpingwang@163.com}
}
%
\maketitle

\begingroup
\renewcommand\thefootnote{\Letter}
\footnotetext{Corresponding author.\quad 
\textsuperscript{\dag} Equal contribution.\quad
\textsuperscript{\ddag} This work was carried out during an internship at Qiji Technology.\quad
\textsuperscript{*} This work is supported by the Shenzhen Science and Technology Program under Grant KJZD20240903103702004.}
\endgroup

\begin{abstract}
Despite the ubiquity of modern face retrieval systems, their retrieval stage is often outsourced to third-party entities, posing significant risks to user portrait privacy. Although homomorphic encryption (HE) offers strong security guarantees by enabling arithmetic computations in the cipher space, its high computational inefficiency makes it unsuitable for real-time, real-world applications. To address this issue, we propose \textbf{\textit{Cancelable Product Quantization}} to achieve efficient and secure face retrieval. Our framework comprises: \textit{\underline{(i)} a high-throughput cancelable PQ indexing module} for fast candidate filtering, and \textit{\underline{(ii)} a fine-grained cipher-space retrieval module} for final precise face ranking. A tailored protection mechanism is designed to secure the indexing module for cancelable biometric authentication while ensuring efficiency. Experiments on benchmark datasets demonstrate that our method achieves an decent balance between effectiveness, efficiency and security.
\end{abstract}


\section{Introduction}\label{sec:intro}
The widespread adoption of face retrieval systems highlights the critical role of facial information in biometric authentication. However, this growing reliance also raises significant security and privacy concerns. In typical public facial authentication systems, a third-party computational service platform is often involved ~\cite{lu2009enabling,tang2024icmr24}, where users' biometric features are transmitted for similarity computations and ranking. While this approach enhances commercial viability, it also introduces considerable security risks, as third-party platforms may not always be fully trustworthy. Recent studies ~\cite{lu2009enabling,pitroda2023ppcbir_survey,mai2018reconstruction} show that raw user images can be reconstructed from latent representations, posing severe risks as facial biometrics often link to financial and confidential data. To prevent potential financial and privacy breaches, ensuring robust security and privacy in biometric authentication systems is imperative.

Although traditional symmetric encryption methods like AES\cite{nist2001aes} can safeguard the confidentiality of outsourced data, their reliance on decryption for similarity comparisons renders them ineffective for scalable retrieval. To address this limitation and enable search directly in the cipher space, research efforts have emerged from \textit{privacy-preserving content-based image retrieval} ({\textit{PPCBIR}}\cite{lu2009enabling,pitroda2023ppcbir_survey}) and \textit{biometric template protection} ({\textit{BTP}})\cite{hahn2022btp_survey}. PPCBIR emphasizes more on secure feature representation and BTP primarily focuses on secure authentication mechanisms.
Existing PPCBIR schemes have explored encrypted visual word frequencies for secure image representation \cite{lu2009enabling}, local sensitive hashing for improved efficiency ~\cite{song2020efficient,tang2024icmr24}, homomorphic encryption for cipher-space similarity calculations \cite{engelsma2022hers}, and DNN-based facial representations for enhanced accuracy \cite{li2020similarity}.
However, the security guarantees among PPCBIR methods are neither consistent nor stringent. In contrast, BTP formally defines two explicit security evaluation aspects for reliable biometric recognition: \textit{{invertibility}} and \textit{{unlinkability}}\cite{hahn2022btp_survey}. To ensure security, various methods have been proposed in the literature, including homomorphic encryption (HE) \cite{engelsma2022hers}, hashing \cite{dong2021secure}, feature transformation \cite{hahn2022towards}, and neural network-based joint learning \cite{lee2021softmaxout}. However, existing BTP approaches have paid less attention to retrieval and indexing mechanisms.

In summary, current methods often fall short in jointly achieving effectiveness, efficiency, and security, which hinders their practical application. Striking a balance among these objectives remains an ongoing open problem. As two representatives, HERS \cite{engelsma2022hers}, a state-of-the-art method in BTP, focuses on accelerating HE to achieve high efficiency while ensuring cryptographic-level security. However, it neglects the index building stage and lacks flexibility with respect to various representation compression methods. Conversely, ELSEIR \cite{tang2024icmr24} employs irreversible hashing and differential privacy-based perturbation for scalable image retrieval. However, its omission of HE results in a lower level of confidentiality than HERS, making it unsuitable for security-critical applications.


To tackle these challenges, we propose \textbf{\textit{Cancelable Product Quantization}}, a highly efficient framework for secure face representation retrieval. Our hierarchical two-stage framework consists of: \textit{\underline{(i)} a high-throughput Cancelable PQ indexing module} for fast candidate filtering and \textit{\underline{(ii)} a fine-grained cipher-space retrieval module} for precise final face ranking. For the former, we design a tailored protection mechanism based on random permutation and projection to safeguard the PQ codebook and distance table. For the latter, we employ Full Homomorphic Encryption (Full-HE) to ensure robust security throughout the retrieval process. Our contributions can be summarized as follows:
\begin{itemize}[leftmargin=*, itemsep=0em, parsep=0em]
    \item We propose a fast two-stage secure face retrieval framework that significantly enhances the practical applicability of encrypted retrieval systems.
    \item We propose a novel Cancelable PQ indexing module substantially reduces computational time while maintaining strong biometric template protection.
    \item Extensive experiments on benchmark face retrieval datasets demonstrate that our method surpasses traditional HE-based retrieval approaches in both effectiveness and efficiency while meeting biometric security requirements.
\end{itemize}

\begin{figure}[t]
    \centering
    \includegraphics[width=\linewidth]{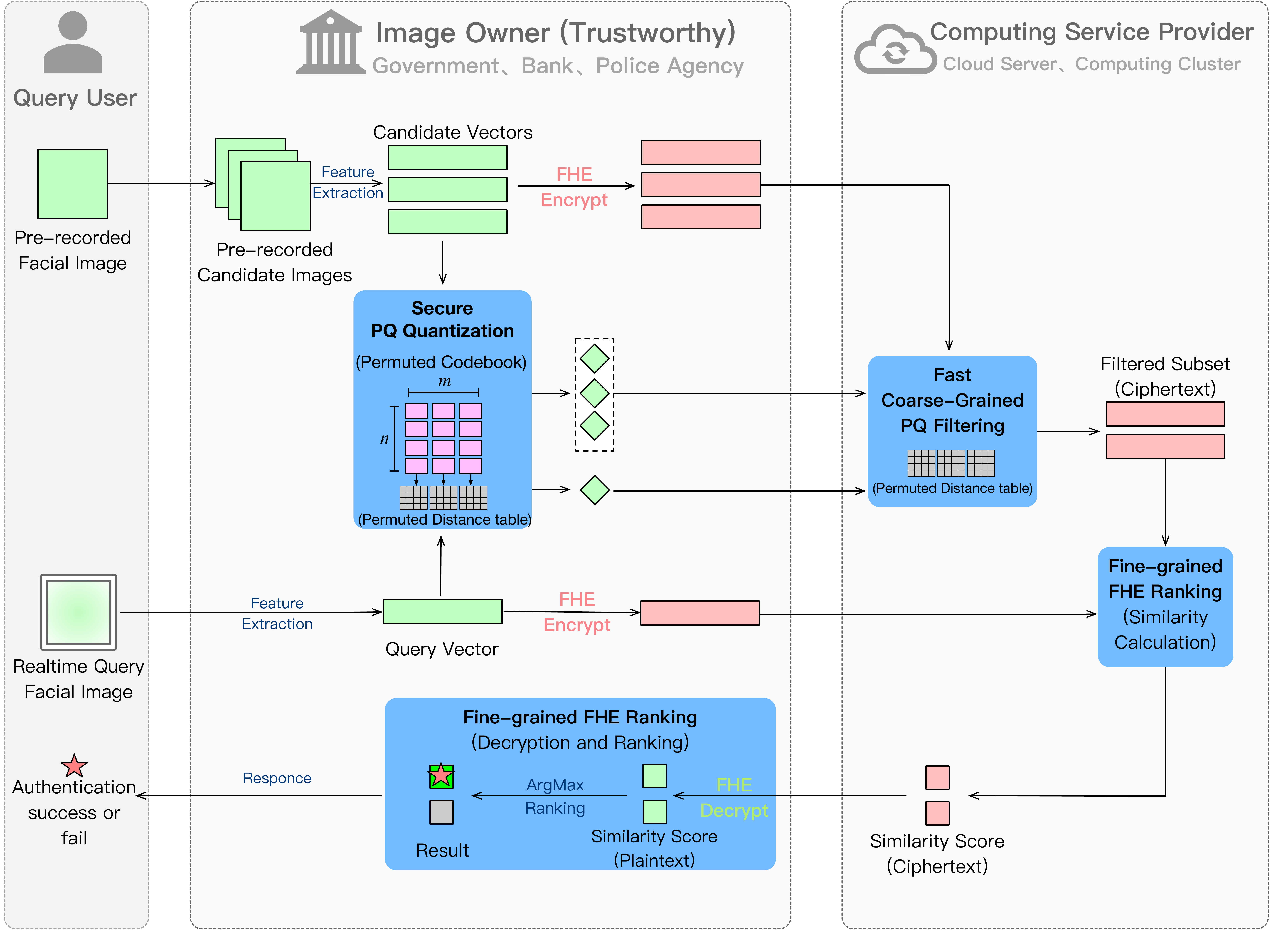}
    \caption{The overall pipeline of our secure retrieval schema.}
    \label{fig:overview}
\end{figure}

\section{Related Work}
\label{sec:rw}

\subsubsection{Privacy-Preserved Image Retrieval}
Privacy-preserving image retrieval \cite{lu2009enabling} aims to support similarity-based search while preventing the leakage of sensitive visual or semantic content. A direct focused on cryptographic primitives, including homomorphic encryption~\cite{gentry2009fully,engelsma2022hers} and secure multi-party computation~\cite{paillier1999public}, to enable secure feature matching over encrypted domains, albeit with substantial computational overhead. To enhance scalability and efficiency, recent approaches incorporate learning-based transformations that preserve retrieval utility while concealing semantic content. Adversarial feature learning~\cite{osadchy2017cnn} and privacy-preserving hashing~\cite{erkin2009privacy,xu2020privacy,song2020efficient,tang2024icmr24} represent key directions, offering robust resistance to inversion and attribute inference attacks. Furthermore, federated learning-based image retrieval systems~\cite{liu2021privacy} and secure indexing mechanisms~\cite{zhang2018privacy} are proposed to facilitate distributed and privacy-aware retrieval pipelines without centralizing raw image data. However, most privacy-preserving image retrieval methods offer unclear guarantees regarding authentication security or suffer from significant cost due to its use of cryptographic operations.

\subsubsection{Biometric Template Protection}
Biometric template protection focuses on securing sensitive biometric data to prevent identity theft, cross-matching, and replay attacks\cite{hahn2022btp_survey}. Foundational methods include biometric cryptosystems, such as fuzzy vaults ~\cite{juels2002fuzzy} and fuzzy commitment~\cite{juels1999fuzzy}, which integrate biometric features with cryptographic constructs. In parallel, cancelable biometrics~\cite{ratha2007generating} provide revocable and non-invertible template transformation to ensure security and diversity. With the emergence of deep learning, new methods leverage representation learning to generate robust and privacy-preserving templates~\cite{pandey2016deep,li2021deep,lee2021softmaxout}. Techniques based on homomorphic encryption~\cite{feng2019fingerprint,engelsma2022hers}, differential privacy~\cite{abadi2016deep}, and adversarial training~\cite{hitaj2017deep} further enhance resilience against inversion and inference attacks. The key research objective remains balancing biometric recognition performance with unlinkability, revocability, and robustness against template leakage. Despite progress in achieving formal security guarantees and effective representation learning, the efficiency of biometric template protection (BTP) methods remains under-optimized and insufficiently studied.

\section{Methodology}
\label{sec:method}

\subsection{Threat Model}
\label{subsec:models}

As shown in Fig\ref{fig:overview}, Our system consists of three key entities: 

\begin{itemize}[leftmargin=*, itemsep=0em, parsep=0em]
    \item \textbf{Query User (QU)}: The user who submits search queries to retrieve images or similar embeddings but does not have direct access to the raw database.
    \item \textbf{Image Owner (IO)}: The entity responsible for uploading and managing an image dataset while ensuring that all stored images and their corresponding embeddings are securely encrypted before indexing.
    \item \textbf{Cloud Service Provider (CSP)}: A third-party service that executes high-speed, parallel retrieval operations in the encrypted domain but is restricted from accessing user information or decrypting stored data.
\end{itemize}

Among these entities, the \textit{Image Owner (IO) is assumed to be fully trusted}, ensuring the security of image storage and encryption. In contrast, both the \textit{Query User (QU) and Cloud Service Provider (CSP) are considered potential adversaries}, as they may attempt to exploit the system for unauthorized benefits. We define two security threats in our system:

\begin{enumerate}[leftmargin=*, itemsep=0em, parsep=0em]
    \item \textbf{\textit{Malicious Query User}}: Attempts to exploit unauthorized biometric features or cryptographic keys to gain illicit access to the system.
    \item \textbf{\textit{Honest-but-Curious Cloud Service Provider}}: Although the CSP adheres to the retrieval protocol, it may attempt to analyze encrypted queries, indices, and stored data to infer biometric representations (e.g., facial features), thereby compromising user privacy.
\end{enumerate}

Our trustworthy retrieval system is designed to mitigate these threats while maintaining high efficiency and accuracy. Specifically, following the security formulations elaborated in ~\cite{maltoni2009handbook,teoh2006randomproj,jin2017indexofmax,pinto2021secure}, our design goals include: \textit{\textbf{(1) Un-invertibility}}: Ensuring that cipher-texted embeddings cannot be reversed to reconstruct original biometric data. \textit{\textbf{(2) Diversity}}: Allowing the protected template to be easily revoked and reissued in case of compromise. \textit{\textbf{(3) Unlinkability}}: Preventing multiple queries from the same user from exhibiting recognizable patterns that could be correlated. \textit{\textbf{(4) Performance}}: Ensuring computationally efficient retrieval without notable accuracy compromise.

\subsection{Proposed Framework}
Our framework is designed from two key aspects: (1) a highly efficient backbone retrieval scheme, {\textit{PQ-indexed Encrypted retrieval}}, and (2) its tailored protection mechanism, {\textit{Cancelable PQ Indexing}}. The former ensures fast and accurate retrieval, while the latter reinforces system security. Specifically, un-invertibility is jointly guaranteed by both components, whereas cancelability is exclusively ensured by the second component. The blue blocks in Figure~\ref{fig:overview} highlight the key modules of our retrieval system.

\subsubsection{\textbf{PQ-Indexed Encrypted Retrieval}}
\label{subsec:fast retrieval schema}
To mitigate the high time complexity of HE-based retrieval, we introduce the three-stage hierarchical framework:

\noindent\textbf{\textit{1) Representation and Compression.}}
We employ the ArcFace network as face encoder, to generate compact and discriminative facial embeddings. To enhance storage and computational efficiency, we apply Principal Component Analysis (PCA) to remove redundant information, improving both efficiency and security. Specifically, given the candidate facial feature set $\mathcal{X}' = \{\mathbf{x}_i' \in \mathbb{R}^{D_1}\}_{i=1}^{N}$, we use PCA to project it into a compressed feature set $\mathcal{X} = \{\mathbf{x}_i \in \mathbb{R}^{D}\}_{i=1}^{N}$, where $D < D_1$. Although we adopt a specific choice in this paper, our method is flexible and can accommodate other feature extractors and compression methods.

\noindent\textbf{\textit{2) Coarse-Grained PQ Filtering.}} 

Given the compressed feature set $\mathcal{X} \in \mathbb{R}^{N \times D}$, we construct a Product Quantization (PQ) index to accelerate approximate nearest neighbor search and filter the original large candidate set into a smaller refined set. This process consists of two main steps: \underline{\textit{(1)}} building the PQ index, including a codebook and a distance lookup table, and \underline{\textit{(2)}} conducting Top-K coarse matching using PQ.

First, we construct the \textit{PQ codebook} $\mathcal{C} = \{\mathcal{C}_i\}_{i=1}^{m}$, where each $\mathcal{C}_i \in \mathbb{R}^{n \times D/m}$ represents a set of $n$ centroids for a subspace of feature vectors. Each vector $\mathbf{x}_i \in \mathcal{X}$ is split into $m$ sub-vectors, and each sub-vector $\mathbf{x}_i^{(j)} \in \mathbb{R}^{D/m}$ is assigned to its closest centroid index, forming the \textit{quantization encoding mapping}:
\begin{align}
    q(\mathbf{x}_i^{(j)}) &= \arg\min_{k} \|\mathbf{x}_i^{(j)} - \mathbf{c}_{j,k} \|^2, \mathbf{c}_{j,k}\in \mathcal{C}_j\\
    Q(\mathbf{x}_i) &= [q(\mathbf{x}_i^{(1)}), q(\mathbf{x}_i^{(2)}), ..., q(\mathbf{x}_i^{(m)})].
\end{align}

Based on the codebook, we pre-compute a \textit{distance lookup table} $\mathcal{D} \in \mathbb{R}^{m \times n \times n}$, where each entry records the squared Euclidean distance between centroids:
\begin{equation}
    \mathcal{D}(j, k_1, k_2) = \|\mathbf{c}_{j,k_1} - \mathbf{c}_{j,k_2} \|^2.
\end{equation}

After constructing the PQ index, we use it to efficiently retrieve a Top-$K$ candidate subset $\mathcal{S}\subset\mathcal{X}$ from the total set of $N$ candidates for each query. Instead of directly computing full vector distances, retrieval is performed by approximating the distance between a query vector $\mathbf{v}$ and a candidate vector $\mathbf{x}_i$ using the lookup table:
\begin{equation}
    d_{PQ}(\mathbf{v}, \mathbf{x}_i) = \sum_{j=1}^{m} \mathcal{D}(j, q(\mathbf{v}^{(j)}), q(\mathbf{x}_i^{(j)})).
\end{equation}
This lookup-based distance computation significantly reduces computational cost while maintaining high retrieval accuracy.

\noindent\textbf{\textit{3) Fine-Grained Homomorphic Ranking.}} The Cloud Service Provider computes encrypted similarity scores for each candidate $\mathbf{x}_k \in \mathcal{S}$ using Fully Homomorphic Encryption (FHE) and sends them to the image owner for decryption:
\begin{align}
    \text{enc}(\text{sim}(\mathbf{v},\mathbf{x}_k)) &= \text{sim}(\text{enc}(\mathbf{v}) \cdot \text{enc}(\mathbf{x}_k))\\
    \text{sim}(\mathbf{v}, \mathbf{x}_k) &= \text{dec}(\text{enc}(\text{sim}(\mathbf{v}, \mathbf{x}_k)))
\end{align}
where $\text{enc}(\cdot)$ and $\text{dec}(\cdot)$ are FHE encryption and decryption, and $\text{sim}(\cdot)$ represents a linear similarity function (e.g., cosine similarity, Euclidean distance). The image owner finally ranks the scores and selects the highest-scoring sample. Such design offers following key advantages:

\begin{itemize}[leftmargin=*, itemsep=0em, parsep=0em]
    \item \textit{Lossless Accuracy:} Fully Homomorphic Encryption (FHE) ensures that similarity computations are carried out without any loss of numerical precision, thereby preserving the fidelity of the retrieval process and guaranteeing that the results remain as accurate as those obtained in the plaintext domain. 
    \item \textit{Strong Security:} By performing all computations in the encrypted space, FHE fundamentally prevents inversion or reconstruction attacks, providing rigorous protection of sensitive data and ensuring that user privacy is maintained throughout the entire retrieval pipeline. 
    \item \textit{Efficiency:} The framework optimizes computational overhead by restricting expensive cryptographic operations to only the filtered candidate set, which substantially reduces the ranking cost while still upholding the strong security guarantees inherent in FHE.
\end{itemize}

\subsubsection{\textbf{Cancelable PQ Indexing}}
\label{subsubsec:cancelable_pq}
To enforce cancelability in the PQ index, we enhance the index structure through random permutation and projection. The process consists of three key phases: \textit{Key Generation}, \textit{Secure Quantization}, and the \textit{Cancelable Retrieval} process.

\noindent\textbf{\textit{1) Key Generation}} Before encoding biometric templates, we generate a key set $k = \{\sigma_1, \dots, \sigma_m, R_1, \dots, R_m\}$, where $\sigma_1, \dots, \sigma_m \in \mathbb{R}^n$ are $m$ independent random permutations, shuffling the PQ codebook to enhance privacy. And $R_1, \dots, R_m \in \mathbb{R}^{n \times D/m}$ are $m$ random projection matrices, transforming sub-vectors into an anonymized space. These transformations obscure the PQ index structure, preventing adversaries from reconstructing the original features.

\noindent\textbf{\textit{2) Secure Quantization.}} To secure feature vectors and the matching process of PQ, we apply two key-bound obfuscation techniques: (1) random projection and (2) codebook permutation:
\begin{align}
    \textcolor{red}{q'}(\mathbf{x}_i^{(j)}) &= \arg\min_{\textcolor{red}{k'}} \|\mathbf{x}_i^{(j)}\textcolor{red}{\mathbf{R}^j} - \mathbf{c}_{j,k'} \|^2, \mathbf{c}_{j,k'}\in \textcolor{red}{\sigma_j}(\mathcal{C}_j) \\
    \mathcal{Q}(\mathbf{x}_i) &= [q'(\mathbf{x}_i^{(1)}), q'(\mathbf{x}_i^{(2)}), ..., q'(\mathbf{x}_i^{(m)})]
\end{align}
Each sub-vector $\mathbf{x}_i^{(j)}$ undergoes a transformation via its respective projection matrix, ensuring randomized feature distributions before indexing. The PQ codebook is then permuted using a key-bound permutation $\sigma_j$, producing a secure quantization mapping. This obfuscation ensures that queries with different keys, even if they are sourced from the same biometric feature, are processed by significantly different quantization codebooks and distance tables, therefore guaranteeing the diversity and unlinkability of protected templates.

\noindent\textbf{\textit{3) Cancelable Retrieval.}} After projection and permutation, retrieval can operate on the protected quantization encoding $\mathcal{Q}(\mathbf{x}_i)$, the permuted codebook $\mathcal{C}'$, and the modified distance table $\mathcal{D}'$. Our design ensures that perturbed retrieval maintains efficiency and effectiveness while offering privacy protection. Below, we justify the advantages in efficiency and cancelability of such design:

\begin{itemize}[leftmargin=*, itemsep=0em, parsep=0em]
    \item \textit{Projection Preserves Distance:} As elaborated in \cite{teoh2006randomproj}, the Jonhnson-Lindenstrauss Lemma \cite{jllemma} guarantees that the random projection using a Gaussian matrix retains pairwise distances, ensuring minimal accuracy loss. The error of this distance preservation is related to the noise ratio within the random projection matrix, as we will discuss in section~\ref{sub:random_proj}.
    \item \textit{Permutation is Accuracy-Lossless:} Codebook permutation alters centroid indexing but preserves pairwise centroid distances. Thus, image owners can compute correctly while cloud service providers (CSPs) remain unable to infer meaningful information from perturbed quantization vectors.
    \item \textit{Secure Encoding Ensures Diversity and Unlinkability}: The combination of random projection and codebook permutation introduces sufficient variation for queries from the same biometric feature. When a certain biometric feature is compromised, the protected template can be simply revoked by applying a different random projection and codebook permutation.
    \item \textit{Permutation strengthens un-invertibility:} The permutation process introduces significant variation in the codebook and the distance table, making it computationally intractable to recover information of the original biometric feature even if one of the codebooks or distance tables are compromised. Specifically, for a quantizer with M subspaces and N codebooks for each subspace, recovering the codebooks via brute force would cost $(N!)^M$ computes, significantly surpassing the computational tractabililty limits.
\end{itemize}


\section{Experiments}
\subsection{Experiment Setting}

\noindent \textbf{Dataset.} To validate the effectiveness of our method, we conduct experiments on well-known face recognition benchmark datasets: \textit{LFW (Labeled Faces in the Wild)}~\cite{huang2007LFW}: The LFW dataset contains 13,233 face images of 5,749 unique individuals, captured in unconstrained environments. It is widely used for evaluating face verification and face recognition algorithms under real-world conditions, given its challenging variations in pose, lighting, and occlusion. \textit{FaceScrub}~\cite{kemelmacher2016megaface}: FaceScrub includes faces of approximately 3.5K celebrities.

\noindent \textbf{Representation Model.} Our model is compatible with arbitrary feature extraction method. In this paper, we apply 512-dimensional embeddings extracted by a publicly available pre-trained \textit{ArcFace} ~\cite{deng2019arcface} model~\footnote{\url{https://github.com/deepinsight/insightface}}.

\noindent \textbf{Baselines.} To evaluate the effectiveness of our proposed methods under a fair confidentiality protection level, we compare them against two representative baselines that operate with the same HE-based security guarantees.: (1) \textit{Vanilla HE} \cite{boddeti2018secure} retrieval method, which performs 1-to-1 ciphertext scoring, decryption, and ranking, and (2) \textit{HERS} \cite{engelsma2022hers}, the state-of-the-art HE accelerated retrieval approach, which significantly improves the efficiency of homomorphic computations. For implementation details, the former is realized using the CKKS scheme with the 1-to-$n$ matching strategy introduced in \cite{boddeti2018secure}. For the latter, we reproduce the method based on its original implementation; however, since the pre-trained DeepMDS++ weights used in HERS are not publicly available, we substitute them with our own self-trained weights for evaluation.

\noindent \textbf{Evaluation Metrics.} We mainly report accuracy and retrieval time for evaluation. All reported accuracy results are Recall@1. For our method, we report the Recall@1 of combined top-5 coarse retrieval and secure fine re-ranking.The retrieval time is reported in average seconds per query. For each experiment, we report the average result of 5 independent runs.

\noindent \textbf{Implementation Details.} Following \cite{boddeti2018secure}, we implement the CKKS Full Homomorphic Encryption using a modified version of the SEAL library, applying a security level of 128 bits for all methods. The training process for Principal Component Analysis (PCA) and Product Quantization (PQ) uses only candidate vectors while excluding query vectors, as in accordance with real-world retrieval scenarios. For our method, the number of subspaces is set to 64 (except for experiment on 32 dimensions, where the number of subspaces is set to 32), and each subspace has a codebook length of 64. A detailed analysis of this parameter selection is provided in the Model Analysis section.

\begin{table}[t]
    \centering
    \caption{Performance on FaceScrub and LFW. We show both the overall retrieval accuracy and average time cost (in seconds) of each query. The results with the best accuracy–efficiency trade-off (ACC - 2 × latency) are highlighted in bold. Our method maintains strong accuracy with runtime reduced from seconds to milliseconds.}
    \label{tab:performance_vertical}
    \resizebox{\textwidth}{!}{
    \begin{tabular}{cc|cccc}
        \toprule
        \textbf{Dataset} & \textbf{Method} & \textbf{32-dim} & \textbf{64-dim} & \textbf{128-dim} & \textbf{256-dim} \\
        \midrule
        \multirow{5}{*}{FaceScrub}
            & FHE               & 92.26 (2.3959)  & 99.25 (4.0878)  & 99.62 (6.7908)  & 99.62 (12.2965) \\
            & HERS              & 69.37 (1.9847)  & 83.18 (3.4525)  & 91.64 (6.3446)  & 96.93 (11.1481) \\
            & Rand Proj Only    & 90.03 (0.0083)  & 92.65 (0.0086)  & 98.77 (0.0094)  & 99.37 (0.0107) \\
            & Permutation Only  & 90.48 (0.0082)  & 93.00 (0.0086)  & 99.15 (0.0093)  & 99.05 (0.0109) \\
            & \underline{\textit{Ours}}     & \textbf{90.81 (0.0083)}  & \textbf{93.50 (0.0088)}  & \textbf{99.62 (0.0096)}  & \textbf{99.62 (0.0110)} \\
        \midrule
        \multirow{5}{*}{LFW}
            & FHE               & 90.12 (0.8913)  & 95.77 (1.7995)  & 95.77 (3.5483)  & 95.77 (6.7806) \\
            & HERS              & 66.28 (0.7814)  & 84.33 (1.4389)  & 90.07 (3.0738)  & 94.43 (6.2139) \\
            & Rand Proj Only    & 89.21 (0.0032)  & 93.70 (0.0056)  & 95.34 (0.0061)  & 95.34 (0.0081) \\
            & Permutation Only  & 88.75 (0.0032)  & 93.24 (0.0054)  & 94.88 (0.0063)  & 94.88 (0.0081) \\
            & \underline{\textit{Ours}}     & \textbf{89.64 (0.0033)}  & \textbf{94.13 (0.0057)}  & \textbf{95.77 (0.0064)}  & \textbf{95.77 (0.0085)} \\
        \bottomrule
    \end{tabular}}
\end{table}

\begin{figure}[ht]
    \centering
    \includegraphics[width=0.85\linewidth]{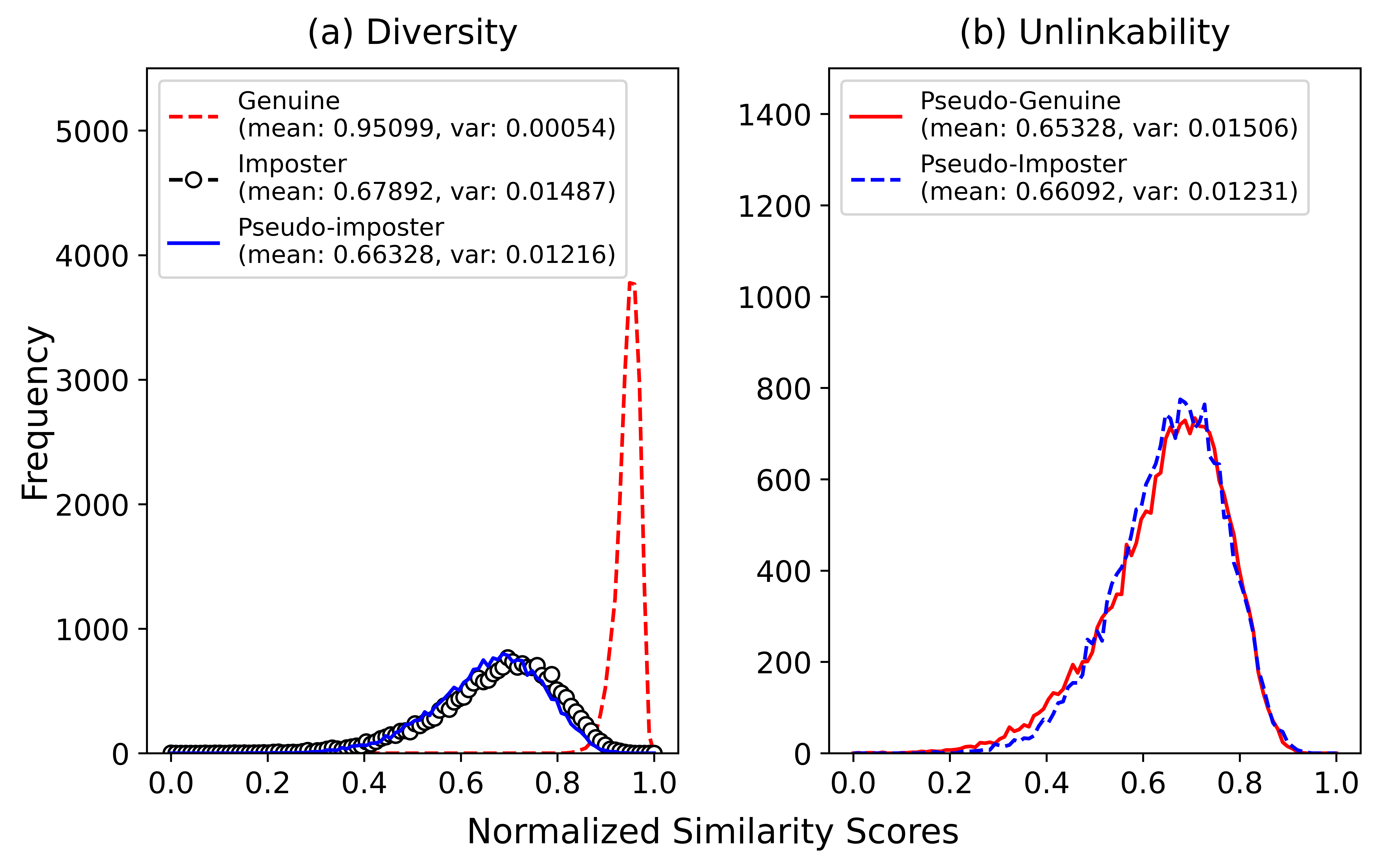}
    \caption{The distribution of similarity scores between different biometric template and random key combinations, demonstrating the diversity and unlinkability properties.}
    \label{fig:dist}
\end{figure}

\subsection{Main Results}
In this section, we present the results on both performance and security, following the security criteria defined in Section~\ref{subsec:models}. Since the property of un-invertibility has already been addressed in Section~\ref{subsubsec:cancelable_pq}, our analysis here focuses specifically on diversity and unlinkability.

\noindent\textbf{Performance} In Table~\ref{tab:performance_vertical}, we report the top-1 recall rate (\%) along with the average retrieval time per query (seconds). Our method achieves 566.30$\times$ acceleration compared to conventional Full Homomorphic Encryption (FHE) and 502.28$\times$ acceleration compared to HERS, while keeping comparable accuracy with FHE and significant accuracy gain compared to HERS, showing advantage in both accuracy and efficiency.

\noindent\textbf{Diversity (Renewability)} 
We prove that the random key encryption process effectively obfuscates biometric similarity by comparing the distributions of \textit{Genuine Scores}, which indicates the similarity between features extracted from the same identity, and \textit{Psedo-Imposter Scores}, which indicates the similarity between these features when encrypted with different keys. As is shown in Figure ~\ref{fig:dist}(a), a significant gap between the pseudo score and genuine score distributions indicates that a biometric feature combined with different random keys produces significantly different protected templates.

\noindent\textbf{Unlinkability} We prove that encrypted templates of the same identity are indistinguishable from those of different identities. To show this, we compare the distributions of \textit{Pseudo-Genuine Scores}, which measure the similarity between encrypted templates of the \textit{same identity} under different encryption keys, and \textit{Pseudo-Imposter Scores}, which measure the similarity between encrypted templates of \textit{different identities} under different encryption keys. As is shown in Figure~\ref{fig:dist}(b), these distributions being similar indicates that multiple encoded versions of the same biometric data are statistically equivalent to multiple encoded versions of different biometric data, therefore guaranteeing unlinkability. Unlinkability prevents multiple encrypted templates generated from the same biometric identity from being linked together. This property is crucial for biometric privacy, as it ensures that attackers who gain access to multiple encrypted templates cannot determine whether they originate from the same identity.

\begin{table}[t]
    \centering
    \caption{The effect of hyper-parameter \(K\) (Avg. 5 runs)}
    \label{tab:K_effect}
    \resizebox{0.85 \linewidth}{!}{
    \begin{tabular}{c|cc|cc}
    \toprule
         \multirow{2}{*}{\textbf{\(K\)}} & \multicolumn{2}{c|}{\textbf{FaceScrub}} & \multicolumn{2}{c}{\textbf{LFW}} \\ 
         \cline{2-5}
         & \textbf{ACC(\%)} & \textbf{Time/Query(ms)} & \textbf{ACC(\%)} & \textbf{Time/Query(ms)} \\ \hline
         1  & 91.75  & 87.08 & 91.05  & 50.05  \\ 
         2  & 92.44  & 87.76  & 92.17  & 50.99 \\ 
         5 & 93.50  & 88.47  & 94.13  & 57.18 \\ 
         10 & 93.66  & 103.74  & 94.35  & 63.23 \\
         100 & 93.71  & 244.15  & 94.39  & 204.20 \\
     \bottomrule
    \end{tabular}
    }
\end{table}

\begin{table}[t]
    \centering
    \caption{The effect of hyper-parameter \(M\) (Avg. 5 runs)}
    \label{tab:M_effect}
    \resizebox{0.85\linewidth}{!}{
    \begin{tabular}{c | c c | c c}
    \toprule
         \multirow{2}{*}{\textbf{\(M\)}} & \multicolumn{2}{c|}{\textbf{FaceScrub}} & \multicolumn{2}{c}{\textbf{LFW}} \\ 
         \cline{2-5}
         & \textbf{ACC(\%)} & \textbf{Time/Query(ms)} & \textbf{ACC(\%)} & \textbf{Time/Query(ms)} \\ \hline
         8  & 60.94 & 39.87  & 63.58  & 20.64  \\ 
         16  & 89.42  & 46.75  & 85.46  & 29.18 \\ 
         32 & 93.18  & 63.28  & 91.29  & 45.21 \\ 
         64 & 99.62  & 96.29  & 95.77  & 64.39 \\
         128 & 99.62  & 219.66  & 95.77  & 147.96 \\
     \bottomrule
    \end{tabular}
    }
\end{table}

\subsection{Ablation Study}

\subsubsection{$K$, the neighborhood size for coarse-grained filtering:} A larger K increases the likelihood of successful matching but also raises the computational overhead of homomorphic encryption (HE) operations during re-ranking. We explore this trade-off in table \ref{tab:K_effect}, under the setting of $D=64, M=64$. We notice that $K=5$ is a good trade-off as further increasing $K$ significantly increases retrieval time but only brings marginal accuracy gain.

\vspace{-3mm}
\begin{table}[t]
    \centering
    \caption{Effect of the perturbation strength of random projection. The parameter $\sigma_{\text{proj}}$ controls the injected noise level. On the right-hand side, $(\mu,\sigma)$ pairs report the results for diversity and unlinkability, while ACC denotes the retrieval accuracy. Our method maintains consistent performance in terms of both security and retrieval effectiveness.}
    \label{tab:sigma_effect}
    \resizebox{\linewidth}{!}{
    \begin{tabular}{cc|ccccccc}
        \toprule
         & $\sigma_{proj}$ & $\mu_g$ & $\mu_i$ & $\mu_{pi}$ & $\sigma^2_g$ & $\sigma^2_i$ & $\sigma^2_{pi}$ & ACC (\%)\\
        \midrule
        Permutation-Only & - & 0.939 & 0.640 & 0.621 & 0.00041 & 0.01383 & 0.01186 & 94.88 \\
        \midrule
        \multirow{4}{*}{Rand. Proj.-$\sigma$} 
        & 1e-4  & 0.951 & 0.669 & 0.675 & 0.00052 & 0.01283  & 0.01404 & 95.15 \\
        & 1e-3  & 0.948 & 0.643 & 0.652 & 0.00051 & 0.01329 & 0.01427 & 95.69 \\
        & 2e-3 & 0.944 & 0.642 & 0.645 & 0.00054 & 0.01488 & 0.01475 & 95.74 \\
        & 5e-3 & 0.947 & 0.639 & 0.648 & 0.00061 & 0.01457 & 0.01584 & 94.78 \\
        & 1e-2 & 0.828 & 0.632 & 0.644 & 0.00065 & 0.01521 & 0.01647 & 93.56 \\
        \bottomrule
    \end{tabular}
}
\end{table}

\begin{table}[t]
    \centering
    \caption{Time per Query (seconds) on Different Scales.}
    \resizebox{0.8\linewidth}{!}{
    \begin{tabular}{lccccc}
        \toprule
        Database Size & 10k & 100k & 1M & 10M & 100M \\
        \midrule
        HERS & 2.8465 & 34.1897 & 518.2925 & 7654.4189 & 13441.7528 \\
        Ours & 0.0058 & 0.0149 & 0.1875 & 0.4136 & 0.7129 \\
        \bottomrule
    \end{tabular}
    }
    \label{tab:scalability}
\end{table}

\subsubsection{$M$, the number of sub-vectors during quantization:} A larger $M$ improves feature representation but also increases the complexity of distance calculations during the PQ filtering process. We explore this trade-off in table \ref{tab:M_effect}, under the setting of $D=128, K=5$. We find that $M=64$ is a good trade-off: further decreasing $M$ would cause significant deficiency on accuracy, while further increasing $M$ would greatly hinder efficiency.

\subsubsection{$\sigma$, the variation of random projection: }\label{sub:random_proj} Random projection matrices are generated based on Gaussian distributions. A larger $\sigma_{proj}$ brings more radical jittering to the random projection process, which brings better diversity and unlinkability properties, but introduces more noise in the data and hinders accuracy. We explore this trade-off in table \ref{tab:sigma_effect}, under the settings $D=128, M=64, K=5$. The genuine ($\mu_g, \sigma^2_g$) and pseudo-imposter ($\mu_{pi}, \sigma^2_{pi}$) distributions being distinct indicates a better diversity, while the imposter ($\mu_i, \sigma^2_i$) being similar indicates a better unlinkability. We find that $\sigma_{proj}=2e-3$ is a good trade-off.

\subsubsection{The effectiveness of Cipher-space Re-ranking}
To validate the effectiveness of Cipher-space Re-ranking, we compare the top-1 retrieval accuracy and time between retrievals with and without reranking on the LFW dataset, with settings $D=128, M=64, K=5$. As is shown in table~\ref{tab:re-ranking}, re-ranking brings considerable accuracy gain while introducing only minor time overhead.

\begin{table}[h]
    \centering
    \caption{Effectiveness of Cipher-space Re-ranking on FaceScrub and LFW Datasets}
    \resizebox{0.85\linewidth}{!}{
    \begin{tabular}{lccc}
        \toprule
        \textbf{Dataset} & \textbf{Method} & \textbf{Top-1 ACC(\%)} & \textbf{Time/Query(ms)} \\
        \midrule
        \multirow{2}{*}{FaceScrub} & w./o. Re-ranking & 98.49 & 81.75 \\
                                   & w./ Re-ranking & 99.62 & 96.29 \\
        \midrule
        \multirow{2}{*}{LFW}       & w./o. Re-ranking & 94.13 & 49.88 \\
                                   & w./ Re-ranking & 95.77 & 64.39 \\
        \bottomrule
    \end{tabular}
    }
    \label{tab:re-ranking}
\end{table}

\subsubsection{Scalability of our Method}
Our model has shown superior efficiency on public human facial image datasets at the scale of 10k, we explore if this advantage extends to even larger scaled database. We generate dummy vectors and compare the time performance of our method and HERS against 100k, 1M, 10M, and 100M scale vector databases, under the settings of $D=128, K=5, M=64$.
As is shown in table~\ref{tab:scalability}, the retrieval time of our method is scalable on very large vector databases, since the time consumption of PQ index search is insensitive to database size and the scale of HE re-ranking process is limited due to our high-throughput k-NN filtering process.

\section{Conclusion}
In this work, we introduced Cancelable Product Quantization, a novel and efficient framework for secure face representation retrieval. By leveraging a two-stage hierarchical design and a carefully crafted protection mechanism, our method provides a practical solution that balances efficiency and security. Extensive experiments on benchmark datasets demonstrate its robustness and real-world applicability. Given that efficiency remains a major barrier to deploying privacy-preserving image retrieval, our approach offers a promising direction for advancing practical applications in this field.
%
%

\bibliographystyle{splncs04}
\bibliography{main}

\end{document}